\documentclass[conference]{IEEEtran}
\IEEEoverridecommandlockouts
\usepackage{cite}
\usepackage{amsmath,amssymb,amsfonts}
\usepackage{algorithmic}
\usepackage{graphicx}
\usepackage{textcomp}
\usepackage{xcolor}
\usepackage{tikz}
\usepackage{enumitem}
\usepackage{filecontents}
\usepackage{float}
\usepackage{booktabs}
\usepackage{balance}

\usepackage{makecell}
\usepackage{array}
\usepackage{multirow}
\usepackage{subcaption}
\usepackage{listings}
\usepackage[ruled,linesnumbered]{algorithm2e}
\usepackage[hidelinks]{hyperref}

\def\BibTeX{{\rm B\kern-.05em{\sc i\kern-.025em b}\kern-.08em
    T\kern-.1667em\lower.7ex\hbox{E}\kern-.125emX}}

\newcommand{\system}{SPAgent\xspace}

\newcommand{\Figref}[1]{Fig.~\ref{#1}}       
\newcommand{\Tabref}[1]{Table~\ref{#1}}       
\newcommand{\Algref}[1]{Algorithm~\ref{#1}}     
\newcommand{\Secref}[1]{Section~\ref{#1}}
\begin{document}

\title{Reducing Latency of LLM Search Agent via Speculation-based Algorithm-System Co-Design}

\author{
    Zixiao Huang$^{1,2}$,
    Wen Zeng$^{1}$,
    Tianyu Fu$^{1}$,
    Tengxuan Liu$^{1}$,
    Yizhou Sun$^{1}$,
    Ke Hong$^{1,2}$,
    Xinhao Yang$^{1,2}$, \\
    Chengchun Liu$^{3}$,
    Yan Li$^{3}$,
    Quanlu Zhang$^{2}$,
    Guohao Dai$^{4,2}$,
    Zhenhua Zhu$^{1}$,
    Yu Wang$^{1}$\\
    \\
    $^{1}$Tsinghua University
    $^{2}$Infinigence 
    $^{3}$Lenovo
    $^{4}$Shanghai Jiao Tong University
}


\maketitle

\begin{abstract}
    LLM-based search agents achieve strong performance but suffer from severe latency, as each step requires serialized LLM reasoning followed by action of tool execution. 
    We revisit this bottleneck through the lens of speculation.
    While traditional predict-verify speculation paradigm can break serial execution, its benefit remains limited, as it retains the full original workload and adds extra inference overhead.
    We observe that early agent steps often involve simple evidence-gathering, where correct actions can often be predicted without full reasoning. 
    Building on these observations, we present \system, an algorithm-system co-design framework that expands the role of speculation in search agents to reduce latency.
    Algorithmically, \system introduces a two-phase adaptive speculation mechanism that selectively omits verification when safe.
    System-wise, a two-level scheduler regulates speculative requests based on engine load to ensure speculation remains beneficial.
    We implement \system in real-world systems. 
    Across extensive experimental settings, \system achieves up to $1.65\times$ end-to-end speedup while maintaining same or even achieving higher accuracy, enabling practical deployment of multi-step search agents.
\end{abstract}

\begin{IEEEkeywords}
LLM Agent, Speculation, Scheduling
\end{IEEEkeywords}

\section{Introduction}
Large Language Model (LLM)~\cite{gpt,gemini,deepseek,kimi} based search agents~\cite{search-r1,ReSearch,search-o1,manusearch,resp} have emerged as an effective paradigm for complex information-seeking and reasoning tasks~\cite{hotpotqa,2wikimultihop,triviaqa,musique,browsecomp}.
By combining multi-step reasoning with external tool calls, they can iteratively gather evidence and update their intermediate hypotheses.
This enables them to solve problems that standard single-pass LLM generation often fails to handle. 
However, these advantages come with a notable drawback: search agents typically exhibit much higher end-to-end latency.
In complex search-driven agent scenarios, like Deep Research~\cite{openai-deepresearch}, completing a single task can even take up to 30 minutes.

\begin{figure}[t]
    \centering
    \begin{subfigure}[b]{0.48\linewidth}
        \includegraphics[width=\linewidth]{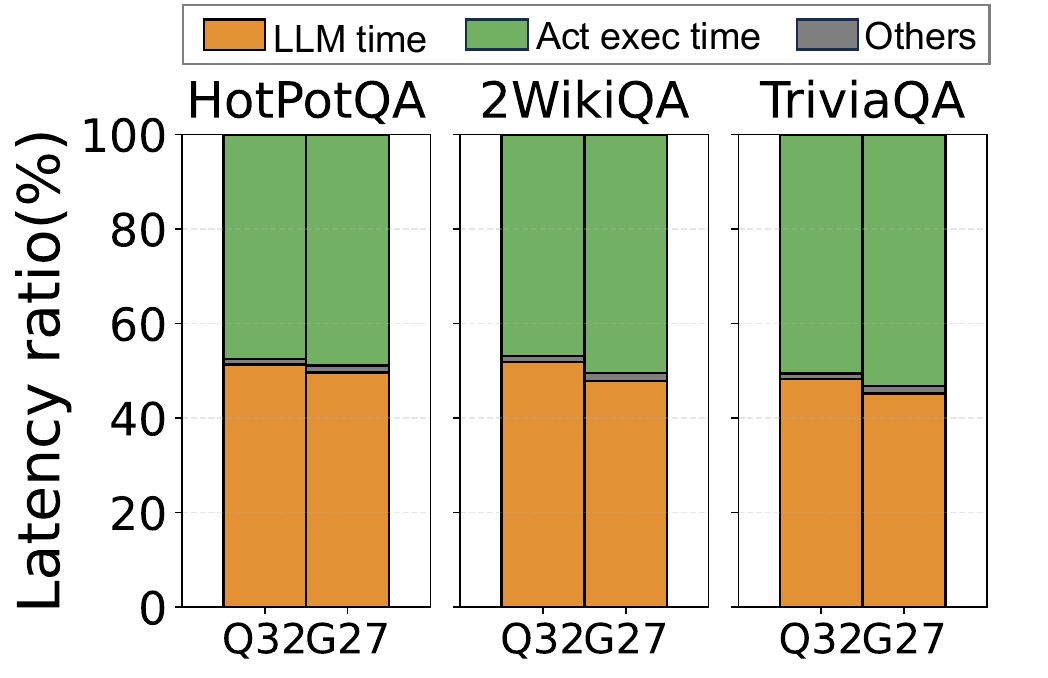}
        \label{fig:latencr_bd_ori}
    \end{subfigure}
    \hfill
    \begin{subfigure}[b]{0.48\linewidth}
        \includegraphics[width=\linewidth]{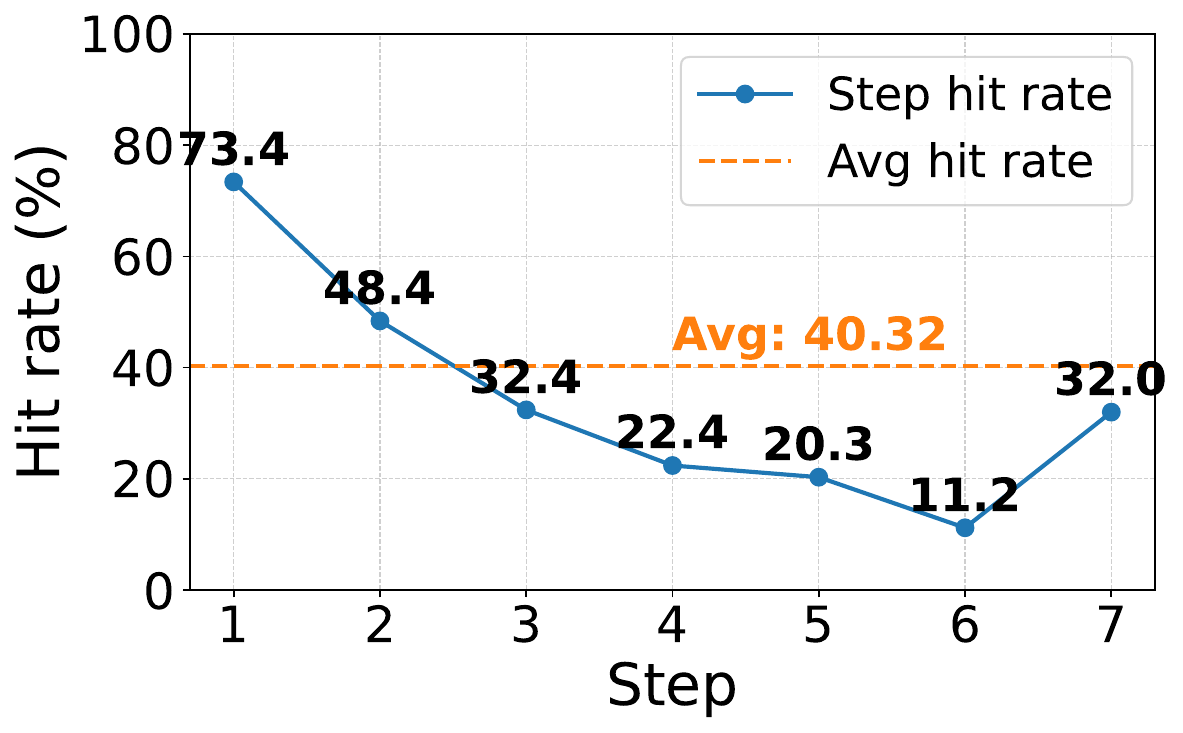}
        \label{fig:hit_rate_step}
    \end{subfigure}
    \vspace{-15pt}
    \caption{(a) Latency breakdown ratio of search agent tasks. (b) Speculative action hit rate across search agent steps.}
    \label{fig:fig1}
    \vspace{-15pt}
\end{figure}

One major source of latency arises from the strict serial dependency in the widely adopted Reason–Action paradigm~\cite{react} of modern LLM search agents.
As illustrated in \Figref{fig:fig1}a, we breakdown the latency of two models (Qwen2.5-32B~\cite{qwen2.5} for Q32, Gemma-3-27B~\cite{Gemma3} for G27) based search agent on three benchmarks~\cite{hotpotqa,2wikimultihop,triviaqa}, both LLM inference and action execution constitute substantial and non-negligible portions of the overall delay.
In each step of the agent workflow, the model must first complete a full inference pass to produce thought and next action; only after the inference finishes can the system execute the action (e.g., issuing a search query via API) and collect the resulting observation.
This rigid ordering forces the two latency-heavy stages to occur strictly one after another, making step-level delay accumulate and ultimately dominate end-to-end task completion time.

To address the long latency stemming from the strict serial dependency, a natural direction is to draw on speculation.
In domains such as branch prediction~\cite{branchpredict} and speculative decoding~\cite{speculativedecoding,speculativesampling,eagle}, a predict-verify paradigm enables future work to be carried out ahead of time and later validated, effectively reducing end-to-end latency.
Following this idea, we propose that LLM search agents can adopt it in the \textbf{action level}: 
Agent can generate actions directly without reasoning tokens, and execute these actions in parallel with the original reasoning progress.

However, the predict-verify paradigm also comes with fundamental limitations: 
it preserves the full original computation while adding speculative work on top, meaning only correct predictions yield speedups, and the additional inference required for speculation introduces non-negligible overhead.

Given these limitations of the predict-verify paradigm, we demonstrate that \textbf{not all speculative actions require verification} in search agents.
As illustrated in \Figref{fig:fig1}b, speculative actions achieve a 73.4\% hit rate in the first step, far exceeding the overall average of 40\%.
Equally important, early-step mistakes are low-risk: 
These actions serve mainly to gather information, and an incorrect speculative action often still task-relevant and sometimes even providing useful auxiliary evidence, rather than causing irreversible deviations in the agent's reasoning trajectory or corrupting internal context~\cite{ircot}.
Thus, omitting verification in early steps offers substantial latency savings by eliminating inference overhead while incurring only minimal, recoverable cost when speculation is wrong.

To reduce the long latency of search agents based on our observations, we propose \system, a holistic algorithm-system co-designed framework that leverages action speculation in search agents that both shorten LLM inference time and overlap tool-execution time.

At the algorithm level, action speculation presents a tension between efficiency and accuracy.
Aggressive speculation without verification can effectively reduce inference latency but risks producing unreasonable actions, whereas speculation with reasoning-based verification preserves accuracy but adds inference overhead.
Thus, an effective agent must adaptively choose between these two modes to minimize latency without harming answer quality (\Secref{sec:algo}).

At the system level, speculation introduces additional inference requests and increases heterogeneity in the workload, as speculative requests produce far shorter outputs than full reasoning ones.
This higher and more uneven load can delay ongoing reasoning and becomes especially problematic under high concurrency~\cite{turbospec}.
Moreover, speculative requests must finish early enough to precede their corresponding reasoning requests; otherwise, they fail to create any overlap for latency reduction.
To address these challenges, \system employs a two-level scheduling strategy (\Secref{sec:system}): an intra-speculation scheduling that adaptively controls and selects speculative launches based on current engine load; and an inter-request scheduling that prioritizes short speculative requests using an Short-Job-First (SJF)~\cite{osconcept} inspired policy to ensure timely completion and effective parallelism.

In summary, this work makes the following contributions:

\begin{itemize}[leftmargin=*]
    \item We introduce a \textbf{new speculation paradigm} for multi-step search agents that goes beyond traditional predict-verify schemes.
    This features a two-phase design that jointly reduces LLM inference latency and overlaps external action-execution latency while preserving answer accuracy.
    \item We design a two-level \textbf{speculation-aware scheduling} architecture that coordinates speculative and other inference requests under real-world inference engine constraints. 
    This system-level co-design ensures that speculation delivers actual latency reduction even in concurrent serving environments.
    \item We build a full system implementation of \system and deployed on three types of GPU platforms (including NVIDIA RTX 3090, A100, A800). We conduct extensive evaluations across diverse models, benchmarks, and deployment scenarios.
    
\end{itemize}
Experiments demonstrate that \system significantly reduces end-to-end latency, achieving up to $1.65\times$ end-to-end speedup while maintaining comparable answer quality, enabling practical deployment of search agents in both cloud and edge scenarios.

\section{Background and Related Works}
\subsection{LLM Search Agent}
\begin{figure}[t]
    \centering
    \includegraphics[width=0.95\linewidth]{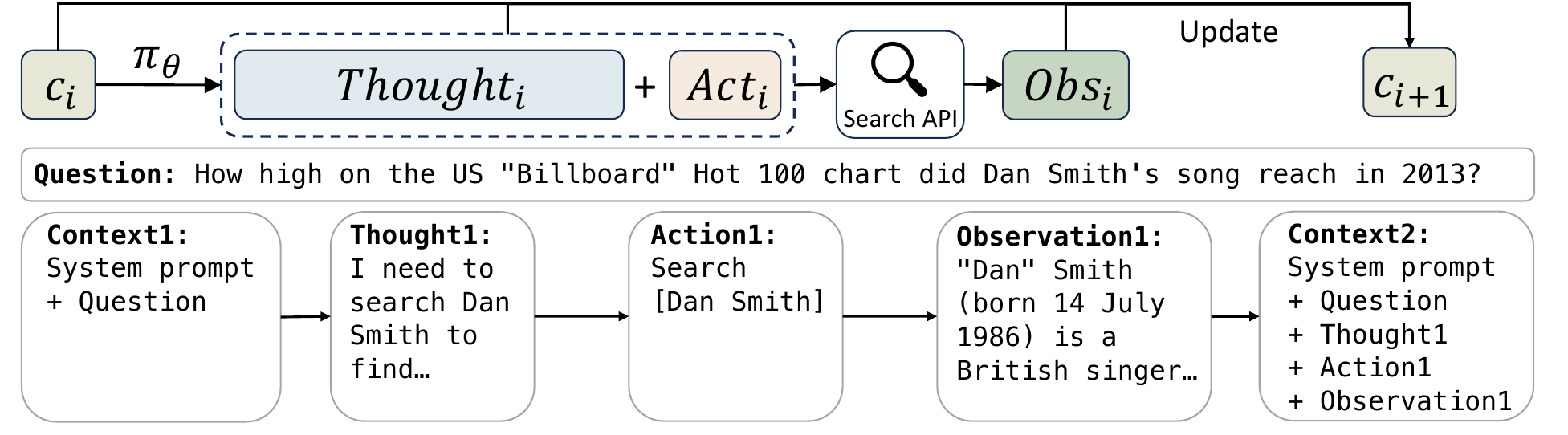}
    \caption{Naive ReAct search agent~\cite{react}}
    \label{fig:naive_step}
\end{figure}

Across the evolution of LLM-based search agents, from workflow-level design~\cite{lats,search-o1} to agents recently trained via reinforcement learning ~\cite{search-r1,ReSearch}, the iterative reason-action has emerged as the de facto standard for evidence collection and decision refinement.
A typical search agent for task solving is shown in \Figref{fig:naive_step}.
At step $i$, the agent maintains a context $c_i$ containing the system prompt along with all previous thoughts, actions, and observations (result of action execution). 
Given $c_i$, the agent generates both a reasoning content $Thought_i$ and an action $Act_i$ by an inference call ($\pi_\theta$) for the current step. 
The action is then executed via the external tools (e.g., search API), yielding an observation that is appended to the context to form $c_{i+1}$. 
This loop continues until the agent generates a final answer and terminates the task.
Under this paradigm, model inference and action execution form a strictly serial dependency, which is the fundamental cause of the high latency in search agents.

\subsection{Speculative Decoding}
Speculative decoding accelerates LLM inference via a predict–verify pipeline, where draft tokens proposed by a smaller model are verified by a larger one~\cite{speculativedecoding,eagle,speculativesampling}.
This reduces decoding latency, but its benefit to search agents is limited because their end-to-end latency is dominated not only by model inference but also by substantial tool-execution time.
Related ideas are being explored in LLM agents.
Works of speculative planning~\cite{speculativeplanning} and speculative actions~\cite{speculativeactions} also follow the predict-verify paradigm.
However, these approaches introduce extra inference requests, which can yield negative effects under high concurrency~\cite{turbospec}.

\subsection{LLM Inference System}
Due to the inherent cost variation in LLM inference~\cite{inferencesurvey}, standard throughput-oriented policies~\cite{vllm,orca} often suffer from head-of-line blocking, where short requests are delayed by long ones. 
This limitation becomes critical in high-concurrency scenarios involving mixed speculative and reasoning requests, necessitating dynamic scheduling to maintain system responsiveness.

\section{Adaptive Action-Level Speculation}
\label{sec:algo}
Despite the strictly serial dependency within the search agent, we argue that action-level speculation can reduce model inference latency and overlap action-execution time to shrink the overall step latency.
In this section, we present a two-phase speculation framework that adapts to the varying reasoning demands across different steps.
By leveraging speculation differently in these two phases, the framework achieves a balanced trade-off between execution efficiency and task accuracy throughout the search task.
We will introduce the motivation, the two distinct phases, their transition mechanism, and the design of action results reuse mechanism through an Action Server.

\subsection{Motivation}
Our adaptive action-level framework is motivated by the reasoning demands vary in steps of search agents.
As shown in \Figref{fig:fig1}b, we allow the model to directly sample candidate actions without reasoning.
Preliminary results indicate that these directly generated speculative actions can often match those produced after reasoning, particularly in the early steps of the task, where we observe 73.4\% match rate in the first step.
This suggests that in the initial stages, where the agent primarily extracts key entities or keywords for search, the reasoning demand is relatively low, and effective actions can be inferred without deep thought.

However, as the task progresses and the agent must integrate accumulated information to plan accurately, deliberate reasoning becomes essential for generating coherent and correct actions.
As an evidence, the speculative action hit rate falls to as low as 11\% in later steps.
Based on this, we design two specific phases to address these different characteristics:
\begin{itemize}[leftmargin=*]
    \item In the \textbf{Aggressive Speculation Phase} (low reasoning complexity), the agent employs speculation to skip reasoning and directly predict actions, thereby reducing the \textit{inference time} spent on thought generation.
    \item In the \textbf{Verified Speculation Phase} (high reasoning complexity), the agent performs reasoning while concurrently applying speculation to predict and execute likely actions in parallel, thereby overlapping the \textit{action execution time}.
\end{itemize}

\begin{figure}
    \centering
    \includegraphics[width=0.95\linewidth]{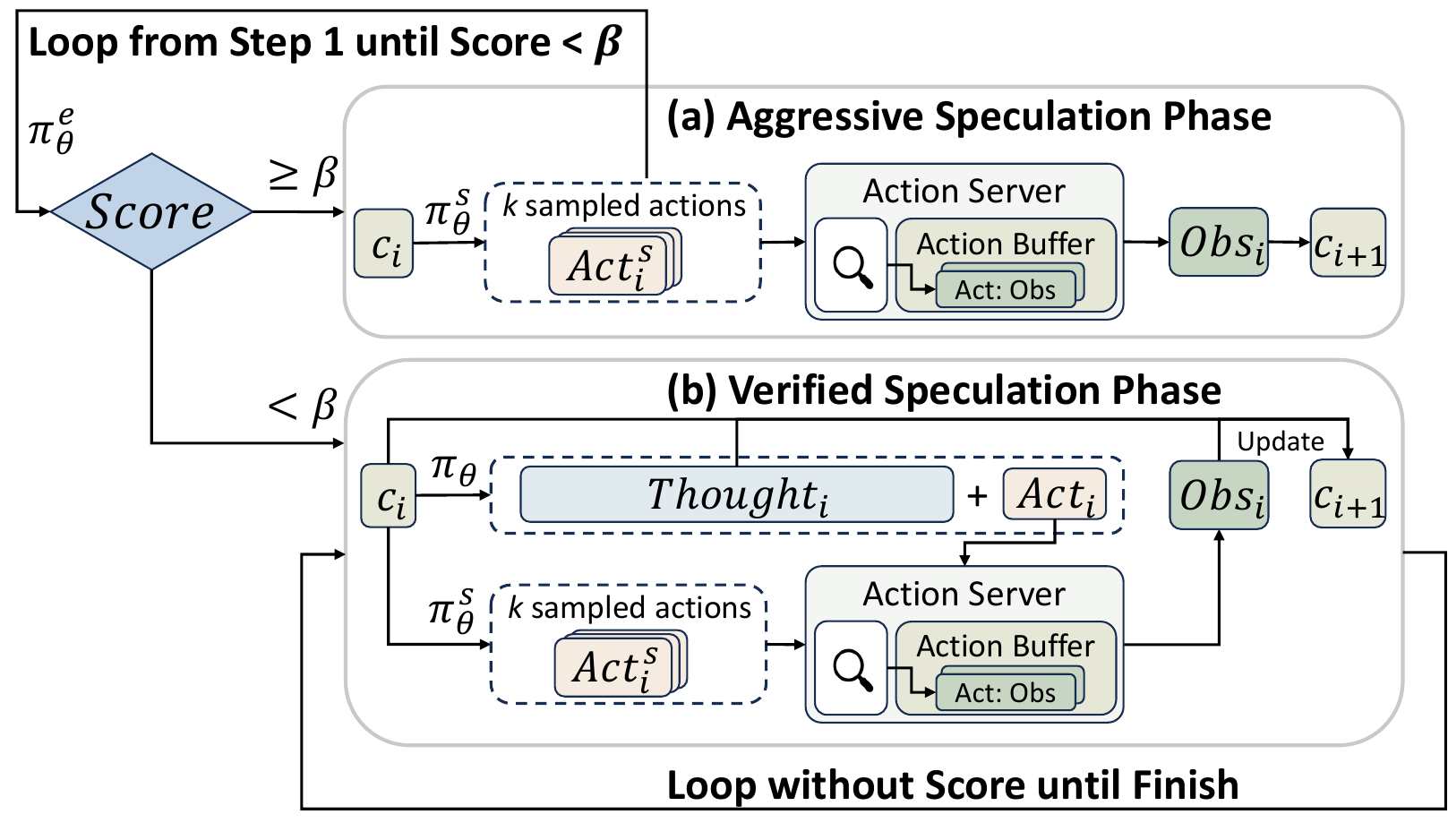}
    \caption{Adaptive Speculation in search agents.}
    \label{fig:ada_spec}
\end{figure}

\subsection{Aggressive Speculation Phase}
\label{sec:aggressive}
This phase appears in the early steps of the search agents, where the reasoning demand is low, direct speculation for actions can get reasonable results.
As shown in \Figref{fig:ada_spec}a, given the current context $c_i$, the agent directly samples $k$ speculative actions through model inference $\pi_{\theta}^s$.
After that, the agent gets the observation result of sampled actions through the Action Server, which contains the search API and Action Buffer to store the action results for speculation. 
We will introduce them in detail in the \Secref{sec:action-server}.
Compared to the original reason-action step in the search agent, the steps in the Aggressive Speculation Phase effectively reduce the latency of reasoning in the model inference.

\subsection{Verified Speculation Phase}
\label{sec:verified}

To complement the Aggressive Speculation Phase and preserve accuracy in reasoning-intensive steps, we introduce the Verified Speculation Phase, where speculation runs alongside normal reasoning.
At each step in \Figref{fig:ada_spec}b, the agent generates the $(Thought, Act)$ pair on the main agent path, while a parallel speculative path samples $k$ candidate actions and executes them immediately.
If a speculative action matches the action later produced in main agent path, its execution result is retrieved directly from the Action Server.
Otherwise, the main path falls back to executing the correct one.

Compared with the original step in the search agent, the steps in Verified Speculation Phase add additional speculation for actions and therefore enable the parallelization between model reasoning in main agent path and action execution in speculative path.
However, this speculation strategy adds extra requests to the model inference system, which may lead to degradation in inference speed, especially in high-concurrency scenarios.
We will tackle this problem in \Secref{sec:intra-sched}.

\begin{figure}[t]
    \centering
    \includegraphics[width=0.5\linewidth]{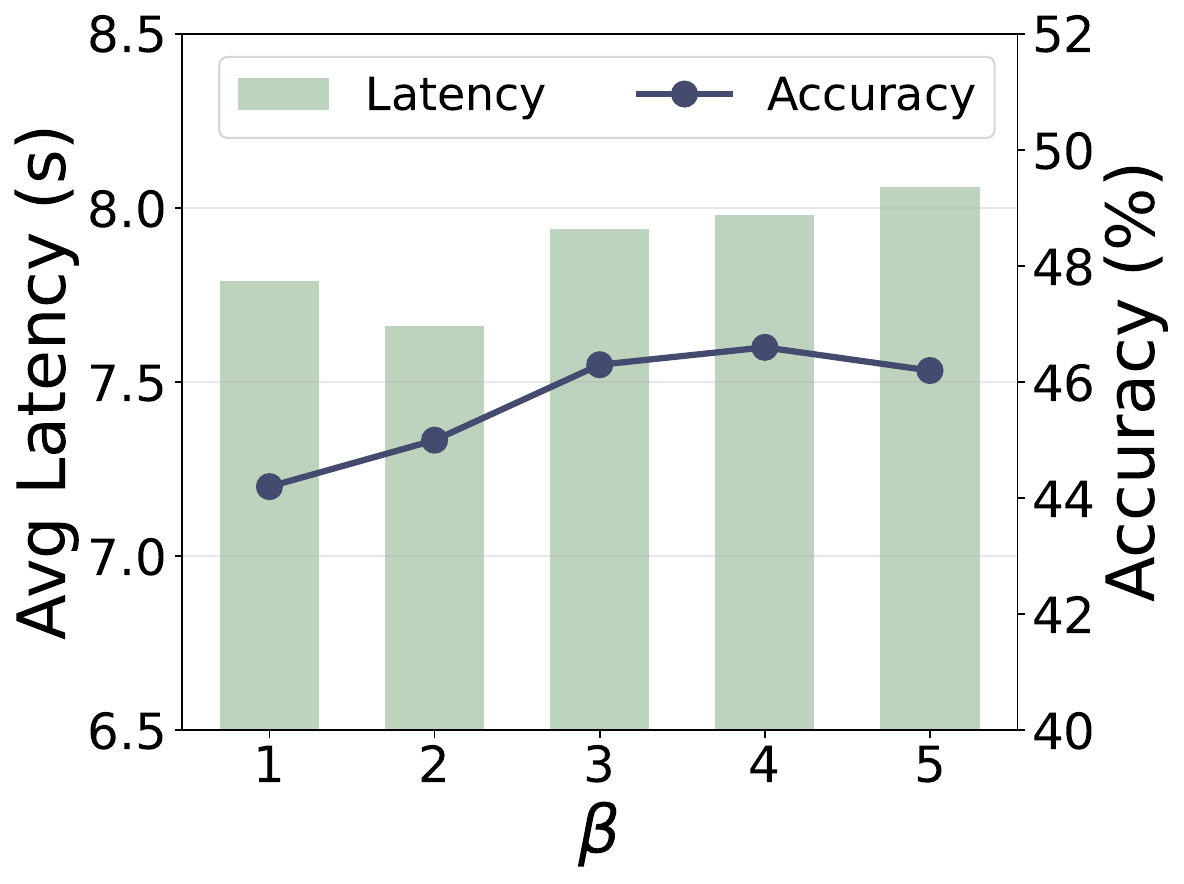}
    \caption{Influence of $\beta$ on accuracy and latency.}
    \label{fig:beta}
\end{figure}

\subsection{Phase Transition}
While each speculation phase offers distinct benefits: 
Aggressive Speculation reduces inference time by omitting reasoning, and Verified Speculation ensures accuracy through reasoning-guided verification and overlaps action execution time. 
Using either mode alone across the entire task is inaccurate or inefficient. 
To adaptively balance efficiency and accuracy, we introduce a dynamic switching mechanism based on the agent's self-reflection~\cite{reflexion}. 
As shown in \Figref{fig:ada_spec} left, after generating speculative actions in the Aggressive Speculation Phase, we invoke an inference call ($\pi_\theta^e$) to score their plausibility.
When all scores of $k$ speculative actions fall below a predefined threshold $\beta$, the system interprets this as a signal that the task has entered a reasoning-intensive stage and transitions to the Verified Speculation Phase. 
This adaptive design allows the agent to skip unnecessary reasoning in the early steps while leveraging verified parallel speculation in later steps, effectively reducing latency without compromising accuracy.

We show the influence of threshold choice in this switching mechanism using Gemma-3-27B on the 2WikiMultiHopQA benchmark, as shown in \Figref{fig:beta}.
Using a 1-5 scoring scale, we observe a clear trade-off: increasing $\beta$ improves answer accuracy but also increases average latency. 
Notably, accuracy gains saturate once $\beta\!>\!3$, while latency continues to grow. 
The results suggest that $\beta\!=\!3$ or $4$ provides the best balance, enabling early-step fast speculation while reliably switching to verified reasoning in complex stages.

\subsection{Action Server}
\label{sec:action-server}
The Action Server is the system component that enables reuse of speculative execution results.
It maintains an in-memory Action Buffer that is implemented as a thread-safe dictionary mapping actions to their execution states and results.
The buffer resides entirely in CPU memory, and its footprint remains small in practice, about 200 Bytes per task.
Incoming action requests are handled concurrently: upon arrival, the server checks the buffer to determine whether the action is already completed or in progress.
If so, the request reuses the stored result or attaches to the ongoing execution, avoiding redundant tool latency.
Only actions absent from the buffer trigger a new search API call.
This design allows the main agent path to immediately consume pre-executed speculative results while ensuring correct synchronization under parallel execution.

\section{Speculation-Aware Scheduling Mechanism}
\label{sec:system}

While the Verified Speculation Phase reduces agent latency by overlapping reasoning and action execution, it introduces additional challenges in online serving scenarios. 
First, the additional speculative requests increase the computational load, potentially degrading overall inference speed. 
Second, inefficient scheduling between speculative and main agent requests can cause speculative action generation to finish later than the corresponding main agent requests. 
Such delays nullify the benefits of parallel execution. 
To address these challenges, we design a two-level scheduling framework: 
\textbf{Intra-Speculation Request Scheduling} dynamically regulates the emission of speculative requests to control inference overhead, and \textbf{Inter-Request Scheduling} globally prioritizes requests to ensure timely speculative completion.

\begin{figure}[t]
    \begin{subfigure}[c]{0.45\linewidth}
        \centering
        \includegraphics[width=\linewidth]{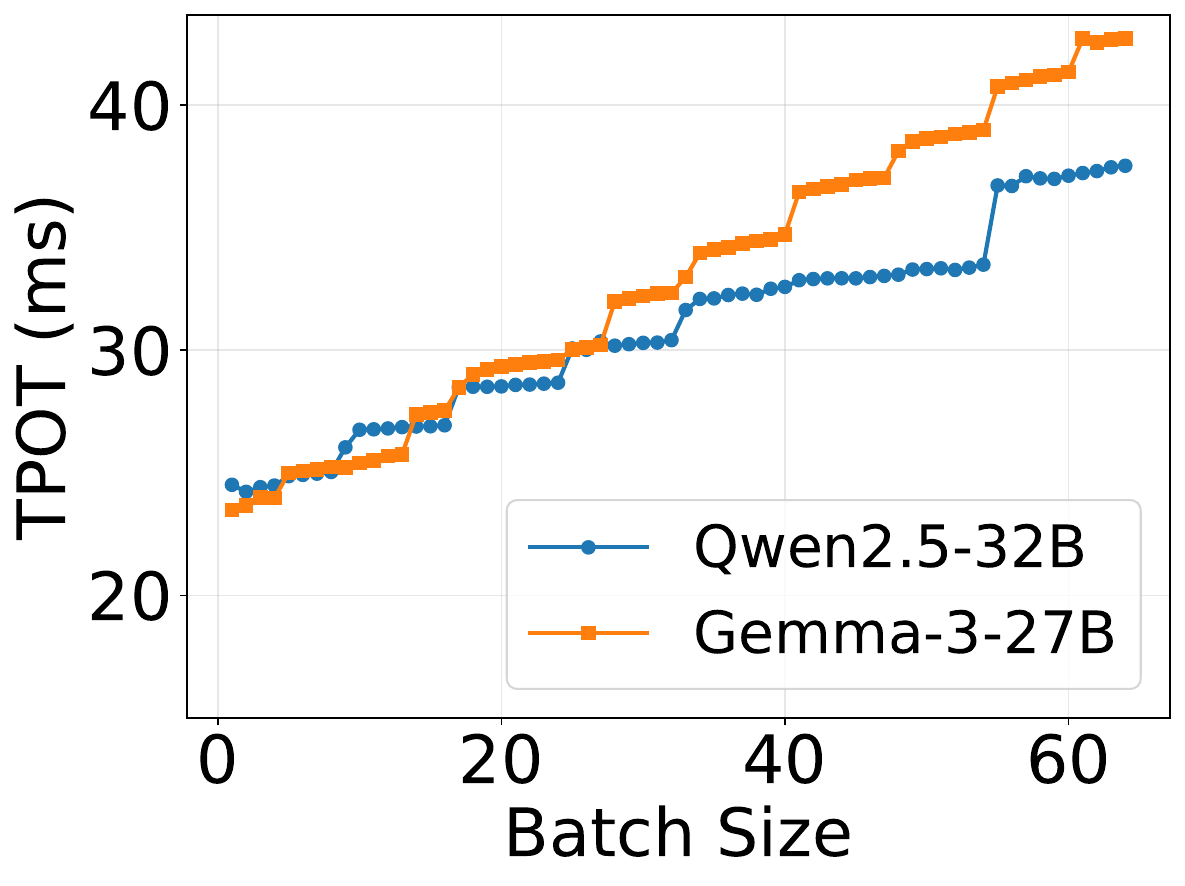}
    \end{subfigure}
    \hfill
    \begin{subfigure}[c]{0.45\linewidth}
        \centering
        \includegraphics[width=\linewidth]{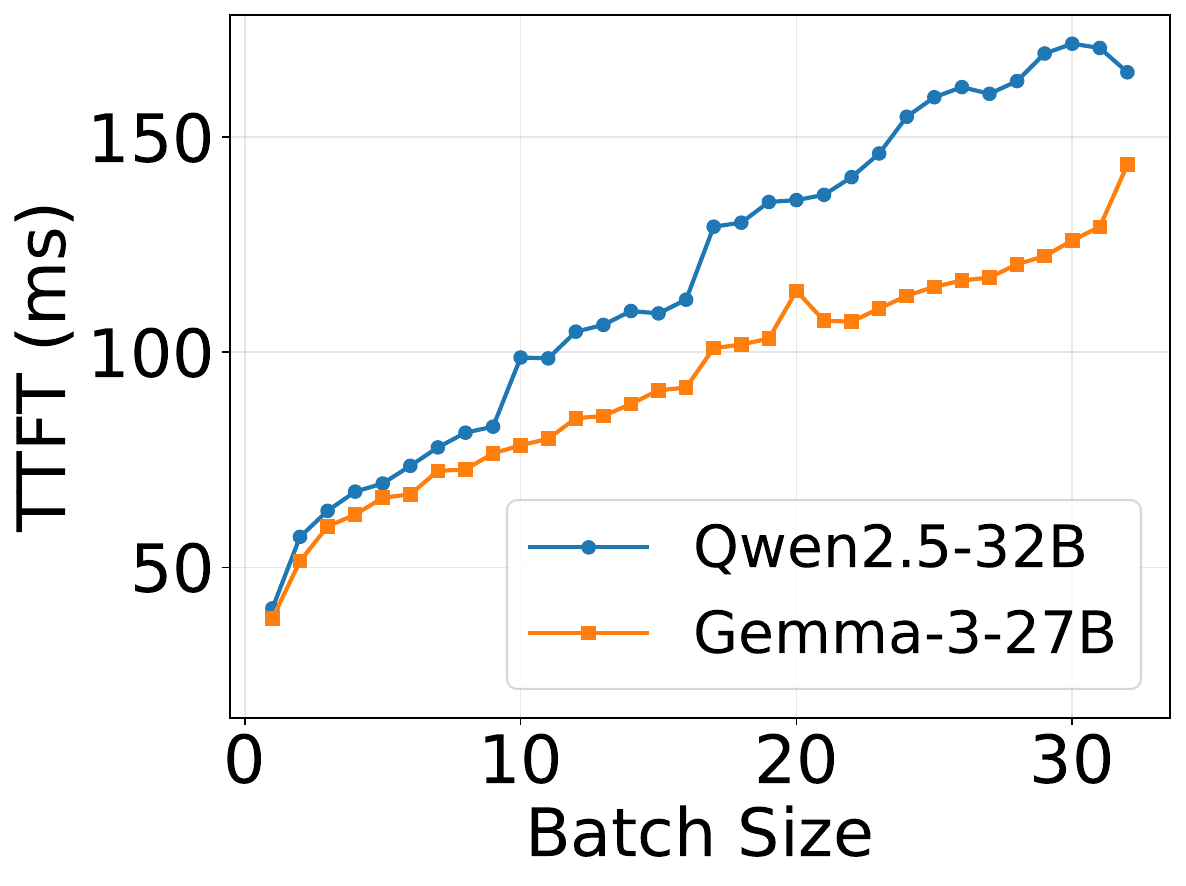}
    \end{subfigure}
    \caption{(a) Average TPOT and (b) Average TTFT on different batch sizes.}
    \label{fig:plots}
\end{figure}

\subsection{Intra-Speculation Request Schedule}
\label{sec:intra-sched}

In the Verified Speculation Phase, speculative actions run concurrently with main agent reasoning to overlap tool latency, but they also introduce extra inference requests. 
Our profiling in \Figref{fig:plots}a and b shows that this overhead is negligible at low concurrency, as decode time remains stable and a single prefill request costs roughly one decode step. 
However, as concurrency increases, speculative requests inflate the number of active decode jobs, slowing per-step decode speed by around to 20\%. 
Moreover, batches begin to contain many prefill requests whose cost scales almost linearly with batch size, making hybrid batches several times more expensive than decode batch, thus blocking other ongoing requests.

These results indicate that speculative requests can easily outweigh their benefits under high load. 
This raises the key question: \emph{which} speculative requests should be launched, and \emph{how many}, so their expected latency reduction exceeds the added inference overhead.
To address this, we introduce the Intra-Speculation Request Schedule, which dynamically selects and emits speculative requests based on real-time engine load.

\begin{table}[t]
    \centering
    \small
    \caption{Notation used in the formulation.}
    \setlength\tabcolsep{2pt} 
    \footnotesize
    \begin{tabular}{ll}
        \toprule
        \textbf{Symbol} & \textbf{Meaning} \\
        \midrule
        $\mathcal{R}$ & Set of running main agent requests \\
        $S$ & Selected subset of $\mathcal{R}$ for speculation \\
        $k$ & Number of speculative samples per request \\
        $N$ & Total number of inference requests ($N_m{+}N_s{+}N_a$) \\
        $N_m$ & \# main agent inference requests ($|\mathcal{R}|$) \\
        $N_s$ & \# speculative inference requests (Verified Speculation phase) \\
        $N_a$ & \# inference requests from Aggressive phase \\
        $t_{act}$ & Average action execution time \\
        $p$ & Probability a speculative sample hits the correct action \\
        $l_s$ & Average speculative output length (tokens) \\
        $L_s$ & Input token length for speculative prefill \\
        $T_h(\cdot)$ & Engine profiling timing results for hybrid batches \\
        $T_{r,a}(S,N)$ & Expected latency reduction from overlap \\
        $T_{o,d}(S,N)$ & Decode phase overhead of speculative requests \\
        $T_{o,p}(S,N)$ & Prefill phase overhead of speculative requests \\
        \bottomrule
    \end{tabular}
    \label{tab:notion}
\end{table}

The symbols used in our scheduling formulation are listed in \Tabref{tab:notion}.
Given the running main agent requests $\mathcal{R}$ in the Verified Speculation Phase, the scheduler selects a subset $S\subseteq\mathcal{R}$ to issue $k$ speculative samples.
The objective is to maximize the expected latency reduction minus inference overhead:
\begin{equation}
    \small
    \max_{S} T_r(S, N) = T_{r,a}(S, N) - (T_{o,d}(S, N)+T_{o,p}(S,N))
\end{equation}
For each $r\in S$, latency is reduced when at least one speculative sample matches the main agent action.
Under independent hit probability $p$, and suppose the action time can be fully overlapped by the main agent path, the expected reduction per task is:
\begin{equation}
    \small
    T_{r,a}(S, N) = \frac{1}{N_m + N_a}\sum_{r\in S} t_{act} \times [1-(1-p)^k]
\end{equation}

The overhead on the inference engine primarily manifests in the prefill and decode phases of speculative requests.
In widely-used engine like vLLM~\cite{vllm}, prefill requests are batched and computed with ongoing decode requests.
Considering the decode phase, speculation increases the number of concurrently processed requests, thereby raising the computational load per decode iteration, resulting in increasing time for one decode step.
Since speculative requests produce very short output length, typically fewer than ten tokens, we assume that the load of inference engine remains stable during each scheduling step.
So the overhead of decode phase is:
\begin{equation}
    \small
    T_{o,d}(S, N) = l_s \times (T_h(\emptyset, N+k\times |S|)-T_h(\emptyset, N))
\end{equation}
where $T_h(\emptyset, N)$ represents the computation time of a hybrid batch with no prefill requests and $N$ decode requests.

Prefill overhead is added once per speculative request, since all samples of one request share the same prefix:
\begin{equation}
    \small
    T_{o, p}(S, N) = T_h((L_s, |S|), N) - T_h(\emptyset, N)
\end{equation}
where $T_h((L_s, |S|), N)$ represents the computation time of a hybrid batch, including $|S|$ prefill requests with $L_s$ input tokens and $N$ decode requests.
\begin{algorithm}[t]
\small
    \caption{Runtime Speculation Selection}
    \label{algo:intra-spec}
    \SetAlgoLined
    \SetKwData{Left}{left}\SetKwData{This}{this}\SetKwData{Up}{up}
    \SetKwFunction{FMain}{Speculation\_Select\_Step}
    \SetKwProg{Fn}{Function}{:}{}
    \SetKwInOut{Input}{Input}\SetKwInOut{Output}{Output}
    \Input{Speculation Priority Queue $Q_r$, Running Requests $N$}
    \Output{Speculation Choice $S$}
    \BlankLine
    \Fn{\FMain{$Q_r, N$}}{
        $S\leftarrow[\:]$\;
        $T_{r, best}\leftarrow 0$\;
        \While{$|Q_r|>0$}{
            $new\_spec\leftarrow Q_r$.pop() \# Candidate with highest priority\;
            \If{$new\_spec$.\text{wait\_time} > $t_w$}{\label{line:discard}
                Continue\;
            }
            $T_{r, cur}\leftarrow T_r(S + new\_spec, N)$\;
            \If{$T_{r, cur}>T_{r, best}$}{\label{line:accept}
                $T_{r, best}\leftarrow T_{r, cur}$\;
                $S$.append($new\_spec$)\;
            }
            \Else{
                Break\label{line:break}\;
            }
        }
        \Return $S$\;
    }
\end{algorithm}

At runtime, the scheduler maintains a priority queue that stores all main agent requests whose speculative actions have not yet been launched.
The priority of candidates is determined by two factors:
\begin{itemize}[leftmargin=*]
    \item First, the step index within the agent task: 
Requests from earlier steps are assigned higher priority, as earlier speculative actions have a higher probability of being adopted by the main agent, thus yielding greater benefit (as discussed in \Secref{sec:algo}).
    \item Second, the arrival time of requests: 
More recently arrived speculative requests are prioritized because they correspond to main agent path inferences that are farther from completion, leaving a longer window for parallel execution between speculative action and ongoing reasoning. 
To preserve this overlap, speculative requests that have waited beyond a predefined time threshold $t_w$ will be discarded, since their results would no longer provide meaningful parallelism or latency savings.
\end{itemize}

Using these two criteria, the scheduler greedily selects speculative requests to maximize the overall expected latency reduction under the current resource constraints of the inference engine.
As shown in \Algref{algo:intra-spec}, during each scheduling step, the scheduler repeatedly pops the highest-priority request from the queue and tentatively adds it to the speculative set $S$, accept it only if doing so increases the expected latency reduction under the current engine load (Line~\ref{line:accept}).
The process stops once the marginal benefit becomes non-positive (Line~\ref{line:break}), after which all selected speculative requests in $S$ are submitted for inference.

\begin{figure}[t]
    \centering
    \includegraphics[width=0.98\linewidth]{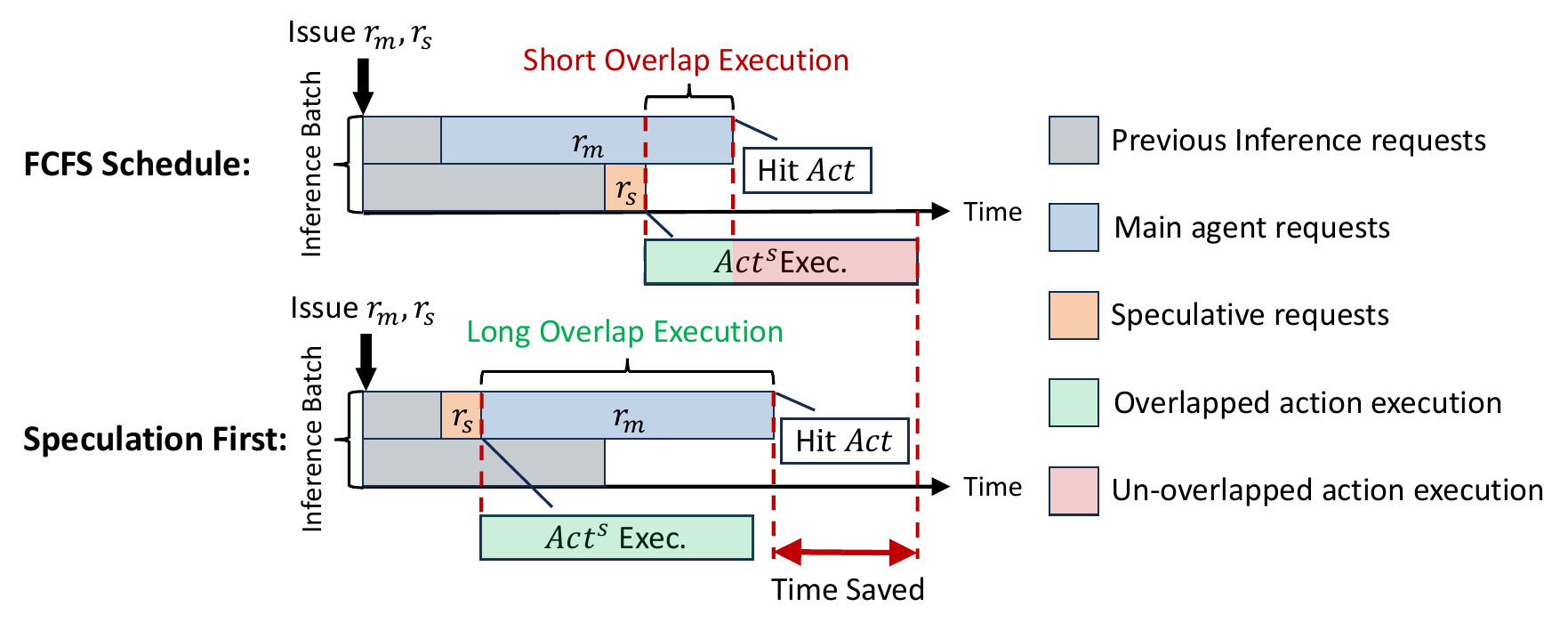}
    \caption{Inter-Request Schedule between speculative requests and main agent requests.}
    \label{fig:schedule}
\end{figure}

\subsection{Inter-Request Schedule}

Even with adaptive emission, speculation remains effective only if speculative requests complete before their corresponding main agent requests; otherwise, their results cannot be reused and the extra inference becomes pure overhead.
However, achieving this timing is non-trivial. 
Speculative and main agent requests have highly asymmetric output lengths: speculation typically produces fewer than ten tokens, while main agent reasoning may produce hundreds.
This asymmetry creates a system-level challenge that under a naive First-Come-First-Serve (FCFS) policy adopted by current inference engine, in which short speculative jobs may frequently queued behind long reasoning jobs, often completing only after they are no longer useful.

\system addresses this challenge through a speculation-first scheduling strategy inspired by Short-Job-First (SJF).
This design is motivated by two observations:
\begin{enumerate}[leftmargin=*]
    \item Speculative requests are extremely short.
    They typically generate fewer than ten tokens, while main agent reasoning may decode hundreds.
    Thus, their end-to-end latency is dominated by queuing delay, not computation.
    Prioritizing them yields disproportionate benefit, since it reduces queueing time without meaningfully delaying longer main agent requests.
    \item Verified speculation is only beneficial if speculative requests finish before its corresponding main agent request.
    In the Verified Speculation Phase, the speculative actions must be available before the main agent reasoning completes; otherwise, no overlap occurs.
    This induces a soft deadline that FCFS cannot satisfy, but SJF-style prioritization naturally does.
\end{enumerate}

As illustrated in \Figref{fig:schedule}, FCFS may cause a speculative request $r_s$ to begin long after its paired main agent request $r_m$, yielding only little overlap.
By contrast, prioritizing speculative requests ensures they begin early, enlarging the overlap window and reducing queuing delay for short jobs.
Even when a speculative action is incorrect, the resulting delay imposed on main agent requests remains minimal due to the tiny output size of speculative requests.

\section{Evaluation}
\subsection{Experimental Setup}
\textbf{Testbed.}
\system is evaluated on three different platforms.
For single-request inference, we evaluate small models on a platform equipped with an AMD EPYC 7H12 64-Core Processor and an NVIDIA RTX 3090 GPU, and evaluate large models on the Intel Xeon Gold 6330 112-core CPU paired with two NVIDIA A100 GPUs. 
For the serving scenario, we experiment on a server configured with an Intel Xeon Platinum 8358 128-core CPU and two NVIDIA A800 GPUs.

\noindent
\textbf{Search Agent Tasks.}
We build our evaluation framework on top of the ReAct agent architecture~\cite{react} (same as concurrent work~\cite{speculativeactions}) and the vLLM inference engine~\cite{vllm}, using the Wikipedia~\cite{wikipedia} API (around 1.5s per request) as the external web-search tool. 
ReAct represents the standardized reason-action workflow in modern search agents, including both workflow-level designs and reinforcement learning–based agents, making it a representative and widely adopted setting for evaluating multi-step reasoning systems. 

To comprehensively assess the performance and generality of \system, we experiment with four different LLMs: Qwen2.5-7B-Instruct (Q7), Qwen2.5-32B-Instruct (Q32)~\cite{qwen2.5}, Gemma-3-4B-it (G4), Gemma-3-27B-it (G27)~\cite{Gemma3}, we set $\text{temperature}=0$ for inference calls in main agent path to ensure result's stability.
We evaluate the agent across three search-oriented benchmarks (HotPotQA~\cite{hotpotqa}, 2WikiMultihopQA~\cite{2wikimultihop}, TriviaQA~\cite{triviaqa}), covering diverse levels of reasoning complexity and interaction depth.
For each benchmark we randomly sample a subset to construct the evaluation suite.

\noindent
\textbf{Baselines.}
We compare \system with representative baselines:
\begin{itemize}[leftmargin=*]
    \item \textbf{Naive ReAct Search Agent + vLLM~\cite{react,vllm}.} This baseline represents the standard ReAct-style search agent using vLLM as the inference engine, without any form of action speculation or system-level scheduling optimization.

    \item \textbf{Speculative Actions~\cite{speculativeactions}.} This baseline introduces speculative action execution using a predict–verify paradigm.  
    Its mechanism is equivalent to the Verified Speculation Phase in our framework. 
    This comparison isolates the benefit of our adaptive speculation strategy and system-level scheduling from previously proposed verified-only speculative action methods.
\end{itemize}
In Speculative Actions and \system, we set the sample number of speculative actions $k=3$ if not specified.

\begin{figure}
    \centering
    \includegraphics[width=0.99\linewidth]{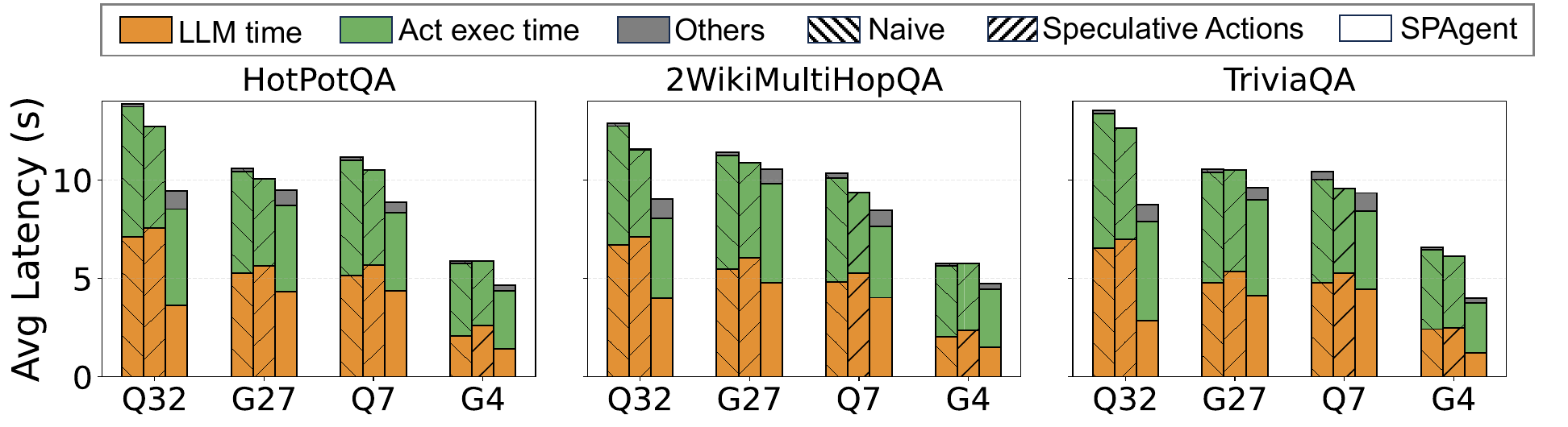}
    \caption{Single-request latency and component breakdown.}
    \label{fig:single}
\end{figure}
\subsection{End-to-End Latency}
\noindent
\textbf{Single Request Inference and Latency Breakdown.}
We compare \system with Naive search agent and Speculative Action across four models and three benchmarks on RTX 3090 (Qwen2.5-7B and Gemma-3-4B) and two A100 GPUs (Qwen2.5-32B and Gemma-3-27B, tensor parallel size $TP=2$).
As shown in \Figref{fig:single}, across all settings, \system achieves $1.08\times$ to $1.65\times$ speed up conpared to the baselines.
In \Figref{fig:single}, LLM time represents for reasoning and action speculation with LLM inference; Act exec time represents the action execution time that is not overlapped with inference; Others in \system is mainly composed of model calls to get action scores.
The improvement arises from reducing both major latency components: the Aggressive Speculation Phase shortens LLM inference by skipping unnecessary reasoning in early simple steps, while the Verified Speculation Phase overlaps action execution with ongoing reasoning.
Overall, \system reduces 23.8\% LLM inference time and 29.4\% un-overlapped action execution time on average, compared with naive agent.
Speculative Actions provides only the latter benefit and increases inference latency by up to 26\% due to extra inference calls, limiting its overall effectiveness.
We also notice that models shows different preference in phase transition.
Qwen models are more likely to speculate aggressively and thus achieves larger inference-time reductions, whereas Gemma models evaluate more conservatively and revert to reasoning sooner, yielding smaller but still consistent gains.

\begin{figure}[t]
    \centering
    \includegraphics[width=0.99\linewidth]{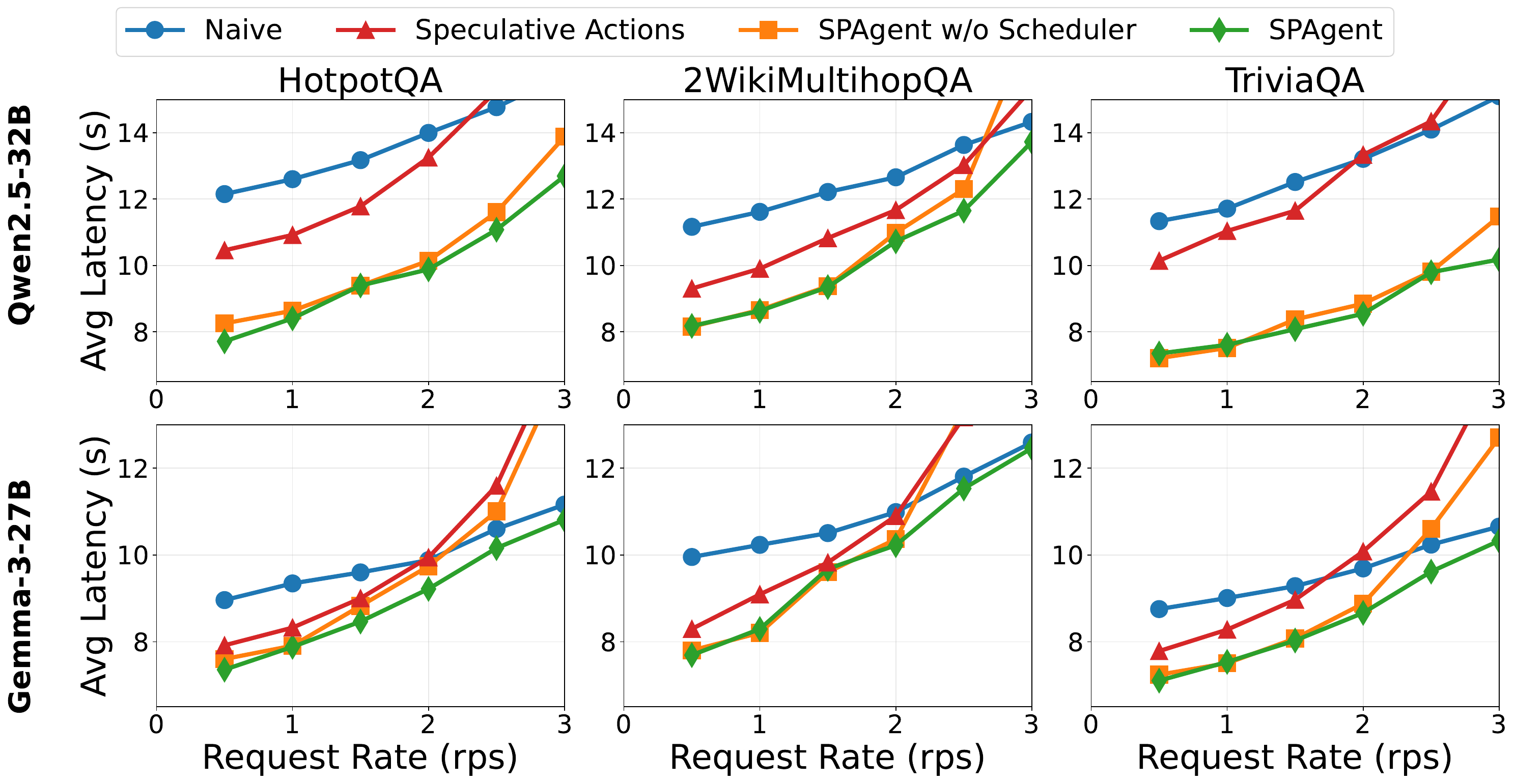}
    \caption{Average latency under different request rates.}
    \label{fig:serve}
\end{figure}

\begin{table}[t]
    \centering
    \caption{Accuracy on different models and benchmarks}
    \setlength\tabcolsep{2pt} 
    \footnotesize
    \resizebox{\linewidth}{!}
    {
    \begin{tabular}{ll|ccc|cc}
        \toprule
        \multicolumn{2}{c|}{Settings}  & HotPotQA & 2WikiQA & TriviaQA & Avg. Acc. & Avg. Speedup\\
        \midrule
        \multirow{2}{*}{Gemma3-4B} & Naive & 26.2 & 25.0 & 38.4 & 29.8 & \multirow{2}{*}{$1.38\times$}\\
        & \system & 25.6 & 24.4 & 41.8 & 30.6\\
        \midrule
        \multirow{2}{*}{Qwen2.5-7B} & Naive & 29.8 & 30.4 & 41.2 & 33.8 & \multirow{2}{*}{$1.19\times$}\\
        & \system & 31.0 & 30.4 & 42.0 & 34.5 \\
        \midrule
        \multirow{2}{*}{Gemma3-27B} & Naive & 45.2 & 46.1 & 62.6 & 51.3 & \multirow{2}{*}{$1.10\times$}\\
        & \system & 44.0 & 46.6 & 63.3 & 51.3 \\
        \midrule
        \multirow{2}{*}{Qwen2.5-32B} & Naive & 41.3 & 41.1 & 63.1 & 48.5 & \multirow{2}{*}{$1.48\times$}\\
        & \system & 42.8 & 42.9 & 68.1 & 51.3\\
        \bottomrule
    \end{tabular}
    }
    \label{tab:accuracy}
\end{table}

\begin{figure*}[t]
    \begin{subfigure}[c]{0.22\linewidth}
        \centering
        \includegraphics[width=\linewidth]{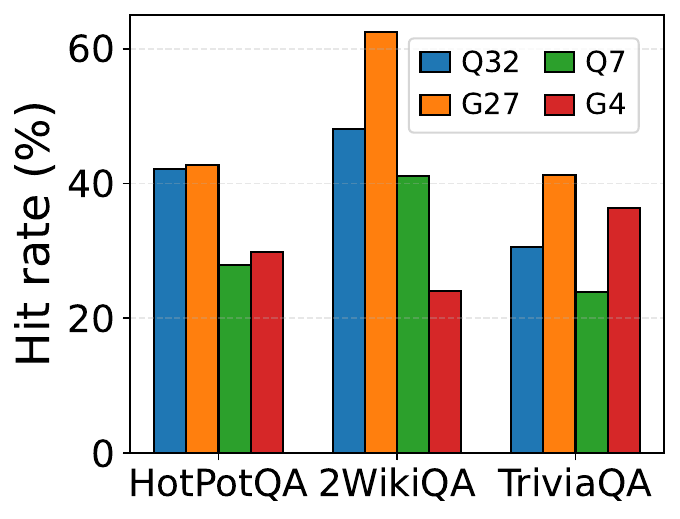}
    \end{subfigure}
    \hfill
    \begin{subfigure}[c]{0.25\linewidth}
        \centering
        \includegraphics[width=\linewidth]{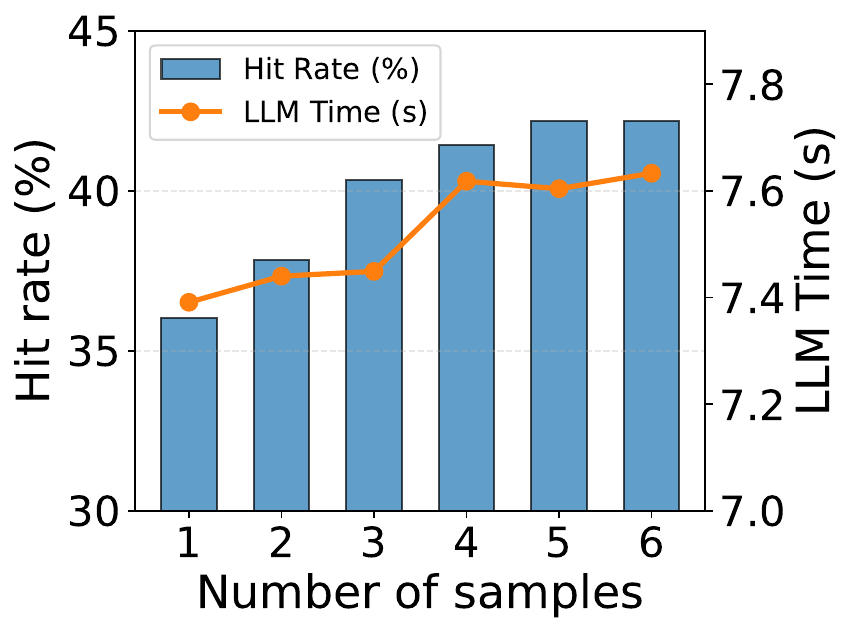}
    \end{subfigure}
    \hfill
    \begin{subfigure}[c]{0.2\linewidth}
        \centering
        \includegraphics[width=\linewidth]{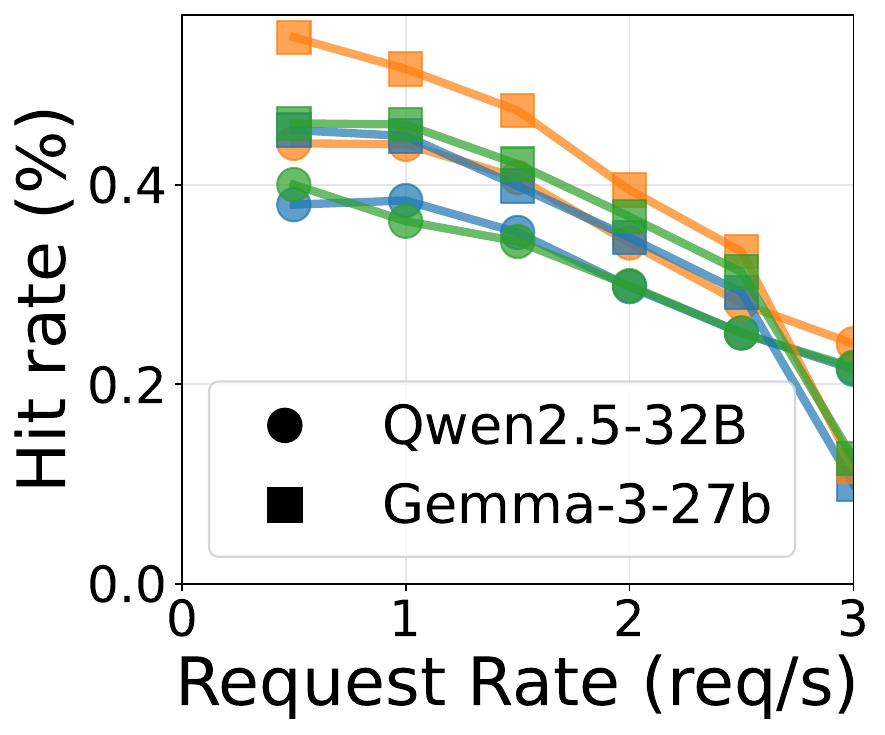}
    \end{subfigure}
    \caption{Action Buffer hit rate in different scenarios. (a) Single-request; (b) Different $k$; (c) Serving scenario.}
    \label{fig:hit_rate_eval}
    \vspace{-10pt}
\end{figure*}

\noindent
\textbf{Online Serving and System Ablation.}
We evaluate \system under online serving scenarios using and Qwen2.5-32B and Gemma-3-27B across three benchmarks on 2 NVIDIA A800 with $TP=2$, comparing against Naive search agent, Speculative Action, and an ablation without scheduling.
We generate search agent tasks' arrivals using a Poisson process, following established methodologies~\cite{vllm,fastserve}.

\Figref{fig:serve} show that \system (with or without scheduler) consistently achieves the lowest mean latency across request rates, speeding up 24.2\% on average, up to 69.6\%, compared to the naive ReAct and Speculative Actions.
Speculative Actions reduce latency compared to Naive only at low concurrency, but becomes up to 49.3\% slower than Naive once load exceeds 2 rps, as its additional inference requests slow down inference steps and overwhelm the small overlap it creates.
The no-scheduler variant suffers from the same issue: speculation increases inference load without any mechanism to regulate it, leading to significant slowdowns under high concurrency.

In contrast, \system remains robust across all loads.
At low request rates, it performs similarly to the no-scheduler variant since speculation is inexpensive. 
As concurrency rises, the scheduler selectively launches speculative requests based on engine load and expected benefit, preventing overload while still exploiting useful overlap. 
Consequently, even at 3 rps, \system avoids the latency spikes of other baselines, demonstrating that speculation is effective only when paired with load-aware scheduling.

\subsection{Accuracy Performance}

Across all evaluated models and benchmarks, \system maintains accuracy on par with the baseline, and in some cases even improves it, as shown in \Tabref{tab:accuracy}.
Notably, the Qwen2.5-32B model exhibit accuracy gains exceeding 5\% on TriviaQA, indicating that the reduced reasoning in the Aggressive Speculation Phase can occasionally enhance decision quality.
A plausible explanation is that skipping early, low-value reasoning may help the model avoid unnecessary or noisy thought chains, thereby preserving focus for later complex steps~\cite{dangeroverthinking,dontoverthinkit,adaptthink}.

The Gemma models show small declines ($\sim 1\%$) on certain benchmarks. 
The modest reductions remain within the typical variance observed when modifying agent prompts or control flows. 
Overall, the results confirm that \system does not introduce meaningful accuracy degradation while delivering substantial latency reductions.

\subsection{Action Buffer Hit Rate}

The Action Buffer hit rate is the proportion of actions generated by the main agent path in the Verified Speculation Phase that can reuse speculative actions already stored in the buffer.
\Figref{fig:hit_rate_eval}a shows Action Buffer hit rate in single-request inference. 
Across models and benchmarks, the hit rate is typically around 40\%, indicating that speculative actions often match the final actions and confirming the practical effectiveness of action-level speculation. 
Note that this metric excludes steps handled by the Aggressive Speculation Phase, which contributes substantially to speedup, so hit rate should not be interpreted as the full extent of speculation’s benefit.
We also evaluate the influence of sample number $k$, shown in \Figref{fig:hit_rate_eval}b.
As the number of sampled speculative actions $k$ increases, the hit rate improves but exhibits diminishing marginal returns, while the additional speculative inference introduces non-negligible overhead. 
Therefore, we set $k=3$ by default, which provides a favorable balance between accuracy gains and computational cost.
In the serving scenarios in \Figref{fig:hit_rate_eval}c (shapes denote models, colors for benchmarks), hit rates naturally decline as request rates increase because the scheduler intentionally suppresses speculative launches under high load to avoid harming inference latency. 
The sharper reduction observed on Gemma model reflects a more conservative scheduling response. 
Overall, these trends demonstrate both the inherent viability of speculative actions and the scheduler’s ability to adapt speculation intensity to system conditions.

\section{Conclusion}
In this work, we introduce \system, an algorithm-system co-design framework that combines adaptive action-level speculation with speculation-aware scheduling to reduce latency of LLM search agents.
Across extensive experiments on models, benchmarks, and deployment scenarios, \system consistently reduces task latency, achieving up to $1.65\times$ end-to-end speedup while preserving accuracy.
Beyond empirical gains, our study broadens the role of speculation in agent design, while insuring the effectiveness of speculation in real-world deployment.
We believe these findings offer a foundation for future advances in agent efficiency research, enabling broader agent-system co-design.

\bibliographystyle{IEEEtran} 
\bibliography{reference}

@article{gpt,
  title={Gpt-4 technical report},
  author={Achiam, Josh and Adler, Steven and Agarwal, Sandhini and Ahmad, Lama and Akkaya, Ilge and Aleman, Florencia Leoni and Almeida, Diogo and Altenschmidt, Janko and Altman, Sam and Anadkat, Shyamal and others},
  journal={arXiv preprint arXiv:2303.08774},
  year={2023}
}

@article{gemini,
  title={Gemini 2.5: Pushing the frontier with advanced reasoning, multimodality, long context, and next generation agentic capabilities},
  author={Comanici, Gheorghe and Bieber, Eric and Schaekermann, Mike and Pasupat, Ice and Sachdeva, Noveen and Dhillon, Inderjit and Blistein, Marcel and Ram, Ori and Zhang, Dan and Rosen, Evan and others},
  journal={arXiv preprint arXiv:2507.06261},
  year={2025}
}

@article{deepseek,
  title={Deepseek-v3 technical report},
  author={Liu, Aixin and Feng, Bei and Xue, Bing and Wang, Bingxuan and Wu, Bochao and Lu, Chengda and Zhao, Chenggang and Deng, Chengqi and Zhang, Chenyu and Ruan, Chong and others},
  journal={arXiv preprint arXiv:2412.19437},
  year={2024}
}

@article{kimi,
  title={Kimi k2: Open agentic intelligence},
  author={Team, Kimi and Bai, Yifan and Bao, Yiping and Chen, Guanduo and Chen, Jiahao and Chen, Ningxin and Chen, Ruijue and Chen, Yanru and Chen, Yuankun and Chen, Yutian and others},
  journal={arXiv preprint arXiv:2507.20534},
  year={2025}
}

@article{search-r1,
  title={Search-r1: Training llms to reason and leverage search engines with reinforcement learning},
  author={Jin, Bowen and Zeng, Hansi and Yue, Zhenrui and Yoon, Jinsung and Arik, Sercan and Wang, Dong and Zamani, Hamed and Han, Jiawei},
  journal={arXiv preprint arXiv:2503.09516},
  year={2025}
}

@article{ReSearch,
  title={Learning to reason with search for llms via reinforcement learning},
  author={Chen, Mingyang and Sun, Linzhuang and Li, Tianpeng and Sun, Haoze and Zhou, Yijie and Zhu, Chenzheng and Wang, Haofen and Pan, Jeff Z and Zhang, Wen and Chen, Huajun and others},
  journal={arXiv preprint arXiv:2503.19470},
  year={2025}
}

@misc{search-o1,
      title={Search-o1: Agentic Search-Enhanced Large Reasoning Models}, 
      author={Xiaoxi Li and Guanting Dong and Jiajie Jin and Yuyao Zhang and Yujia Zhou and Yutao Zhu and Peitian Zhang and Zhicheng Dou},
      year={2025},
      eprint={2501.05366},
      archivePrefix={arXiv},
      primaryClass={cs.AI},
      url={https://arxiv.org/abs/2501.05366}, 
}

@misc{manusearch,
      title={ManuSearch: Democratizing Deep Search in Large Language Models with a Transparent and Open Multi-Agent Framework}, 
      author={Lisheng Huang and Yichen Liu and Jinhao Jiang and Rongxiang Zhang and Jiahao Yan and Junyi Li and Wayne Xin Zhao},
      year={2025},
      eprint={2505.18105},
      archivePrefix={arXiv},
      primaryClass={cs.CL},
      url={https://arxiv.org/abs/2505.18105}, 
}

@inproceedings{resp,
author = {Jiang, Zhouyu and Sun, Mengshu and Liang, Lei and Zhang, Zhiqiang},
title = {Retrieve, Summarize, Plan: Advancing Multi-hop Question Answering with an Iterative Approach},
year = {2025},
isbn = {9798400713316},
publisher = {Association for Computing Machinery},
address = {New York, NY, USA},
url = {https://doi.org/10.1145/3701716.3716889},
doi = {10.1145/3701716.3716889},
booktitle = {Companion Proceedings of the ACM on Web Conference 2025},
pages = {1677–1686},
numpages = {10},
location = {Sydney NSW, Australia},
series = {WWW '25}
}

@misc{fastserve,
      title={Fast Distributed Inference Serving for Large Language Models}, 
      author={Bingyang Wu and Yinmin Zhong and Zili Zhang and Shengyu Liu and Fangyue Liu and Yuanhang Sun and Gang Huang and Xuanzhe Liu and Xin Jin},
      year={2024},
      eprint={2305.05920},
      archivePrefix={arXiv},
      primaryClass={cs.LG},
      url={https://arxiv.org/abs/2305.05920}, 
}

@inproceedings{hotpotqa,
  title={HotpotQA: A dataset for diverse, explainable multi-hop question answering},
  author={Yang, Zhilin and Qi, Peng and Zhang, Saizheng and Bengio, Yoshua and Cohen, William and Salakhutdinov, Ruslan and Manning, Christopher D},
  booktitle={Proceedings of the 2018 conference on empirical methods in natural language processing},
  pages={2369--2380},
  year={2018}
}

@inproceedings{2wikimultihop,
    title = "Constructing A Multi-hop {QA} Dataset for Comprehensive Evaluation of Reasoning Steps",
    author = "Ho, Xanh  and
      Duong Nguyen, Anh-Khoa  and
      Sugawara, Saku  and
      Aizawa, Akiko",
    booktitle = "Proceedings of the 28th International Conference on Computational Linguistics",
    month = dec,
    year = "2020",
    address = "Barcelona, Spain (Online)",
    publisher = "International Committee on Computational Linguistics",
    url = "https://www.aclweb.org/anthology/2020.coling-main.580",
    pages = "6609--6625",
}

@article{triviaqa,
  title={Triviaqa: A large scale distantly supervised challenge dataset for reading comprehension},
  author={Joshi, Mandar and Choi, Eunsol and Weld, Daniel S and Zettlemoyer, Luke},
  journal={arXiv preprint arXiv:1705.03551},
  year={2017}
}

@article{browsecomp,
  title={Browsecomp: A simple yet challenging benchmark for browsing agents},
  author={Wei, Jason and Sun, Zhiqing and Papay, Spencer and McKinney, Scott and Han, Jeffrey and Fulford, Isa and Chung, Hyung Won and Passos, Alex Tachard and Fedus, William and Glaese, Amelia},
  journal={arXiv preprint arXiv:2504.12516},
  year={2025}
}

@misc{musique,
      title={MuSiQue: Multihop Questions via Single-hop Question Composition}, 
      author={Harsh Trivedi and Niranjan Balasubramanian and Tushar Khot and Ashish Sabharwal},
      year={2022},
      eprint={2108.00573},
      archivePrefix={arXiv},
      primaryClass={cs.CL},
      url={https://arxiv.org/abs/2108.00573}, 
}

@misc{openai-deepresearch,
  author       = {OpenAI},
  title        = {Introducing deep research},
  year         = {2025},
  howpublished = {\url{https://openai.com/index/introducing-deep-research/}},
}

@inproceedings{react,
  title={React: Synergizing reasoning and acting in language models},
  author={Yao, Shunyu and Zhao, Jeffrey and Yu, Dian and Du, Nan and Shafran, Izhak and Narasimhan, Karthik R and Cao, Yuan},
  booktitle={The eleventh international conference on learning representations},
  year={2022}
}

@article{reflexion,
  title={Reflexion: Language agents with verbal reinforcement learning},
  author={Shinn, Noah and Cassano, Federico and Gopinath, Ashwin and Narasimhan, Karthik and Yao, Shunyu},
  journal={Advances in Neural Information Processing Systems},
  volume={36},
  pages={8634--8652},
  year={2023}
}

@misc{lats,
      title={Language Agent Tree Search Unifies Reasoning Acting and Planning in Language Models}, 
      author={Andy Zhou and Kai Yan and Michal Shlapentokh-Rothman and Haohan Wang and Yu-Xiong Wang},
      year={2024},
      eprint={2310.04406},
      archivePrefix={arXiv},
      primaryClass={cs.AI},
      url={https://arxiv.org/abs/2310.04406}, 
}

@misc{speculativeactions,
      title={Speculative Actions: A Lossless Framework for Faster Agentic Systems}, 
      author={Naimeng Ye and Arnav Ahuja and Georgios Liargkovas and Yunan Lu and Kostis Kaffes and Tianyi Peng},
      year={2025},
      eprint={2510.04371},
      archivePrefix={arXiv},
      primaryClass={cs.AI},
      url={https://arxiv.org/abs/2510.04371}, 
}

@misc{speculativeplanning,
      title={Interactive Speculative Planning: Enhance Agent Efficiency through Co-design of System and User Interface}, 
      author={Wenyue Hua and Mengting Wan and Shashank Vadrevu and Ryan Nadel and Yongfeng Zhang and Chi Wang},
      year={2024},
      eprint={2410.00079},
      archivePrefix={arXiv},
      primaryClass={cs.MA},
      url={https://arxiv.org/abs/2410.00079}, 
}

@misc{dangeroverthinking,
      title={The Danger of Overthinking: Examining the Reasoning-Action Dilemma in Agentic Tasks}, 
      author={Alejandro Cuadron and Dacheng Li and Wenjie Ma and Xingyao Wang and Yichuan Wang and Siyuan Zhuang and Shu Liu and Luis Gaspar Schroeder and Tian Xia and Huanzhi Mao and Nicholas Thumiger and Aditya Desai and Ion Stoica and Ana Klimovic and Graham Neubig and Joseph E. Gonzalez},
      year={2025},
      eprint={2502.08235},
      archivePrefix={arXiv},
      primaryClass={cs.AI},
      url={https://arxiv.org/abs/2502.08235}, 
}

@misc{turbospec,
      title={TurboSpec: Closed-loop Speculation Control System for Optimizing LLM Serving Goodput}, 
      author={Xiaoxuan Liu and Jongseok Park and Langxiang Hu and Woosuk Kwon and Zhuohan Li and Chen Zhang and Kuntai Du and Xiangxi Mo and Kaichao You and Alvin Cheung and Zhijie Deng and Ion Stoica and Hao Zhang},
      year={2025},
      eprint={2406.14066},
      archivePrefix={arXiv},
      primaryClass={cs.AI},
      url={https://arxiv.org/abs/2406.14066}, 
}

@book{osconcept,
  title={Operating system concepts},
  author={Peterson, James L and Silberschatz, Abraham},
  year={1985},
  publisher={Addison-Wesley Longman Publishing Co., Inc.}
}

@inproceedings{speculativedecoding,
  title={Fast inference from transformers via speculative decoding},
  author={Leviathan, Yaniv and Kalman, Matan and Matias, Yossi},
  booktitle={International Conference on Machine Learning},
  pages={19274--19286},
  year={2023},
  organization={PMLR}
}

@article{speculativesampling,
  title={Accelerating large language model decoding with speculative sampling},
  author={Chen, Charlie and Borgeaud, Sebastian and Irving, Geoffrey and Lespiau, Jean-Baptiste and Sifre, Laurent and Jumper, John},
  journal={arXiv preprint arXiv:2302.01318},
  year={2023}
}

@misc{eagle,
      title={EAGLE: Speculative Sampling Requires Rethinking Feature Uncertainty}, 
      author={Yuhui Li and Fangyun Wei and Chao Zhang and Hongyang Zhang},
      year={2025},
      eprint={2401.15077},
      archivePrefix={arXiv},
      primaryClass={cs.LG},
      url={https://arxiv.org/abs/2401.15077}, 
}

@article{inferencesurvey,
  title={A survey on efficient inference for large language models},
  author={Zhou, Zixuan and Ning, Xuefei and Hong, Ke and Fu, Tianyu and Xu, Jiaming and Li, Shiyao and Lou, Yuming and Wang, Luning and Yuan, Zhihang and Li, Xiuhong and others},
  journal={arXiv preprint arXiv:2404.14294},
  year={2024}
}

@inproceedings{vllm,
  title={Efficient memory management for large language model serving with pagedattention},
  author={Kwon, Woosuk and Li, Zhuohan and Zhuang, Siyuan and Sheng, Ying and Zheng, Lianmin and Yu, Cody Hao and Gonzalez, Joseph and Zhang, Hao and Stoica, Ion},
  booktitle={Proceedings of the 29th symposium on operating systems principles},
  pages={611--626},
  year={2023}
}

@inproceedings {orca,
author = {Gyeong-In Yu and Joo Seong Jeong and Geon-Woo Kim and Soojeong Kim and Byung-Gon Chun},
title = {Orca: A Distributed Serving System for {Transformer-Based} Generative Models},
booktitle = {16th USENIX Symposium on Operating Systems Design and Implementation (OSDI 22)},
year = {2022},
isbn = {978-1-939133-28-1},
address = {Carlsbad, CA},
pages = {521--538},
url = {https://www.usenix.org/conference/osdi22/presentation/yu},
publisher = {USENIX Association},
month = jul
}

@misc{wikipedia,
  title        = {Wikipedia},
  author = "Wikipedia",
  howpublished = {\url{https://www.wikipedia.org/}},
}

@misc{qwen2.5,
    title = {Qwen2.5: A Party of Foundation Models},
    url = {https://qwenlm.github.io/blog/qwen2.5/},
    author = {Qwen Team},
    month = {September},
    year = {2024}
}

@article{Gemma3,
    title={Gemma 3},
    url={https://goo.gle/Gemma3Report},
    publisher={Kaggle},
    author={Gemma Team},
    year={2025}
}

@misc{dontoverthinkit,
      title={Don't Overthink it. Preferring Shorter Thinking Chains for Improved LLM Reasoning}, 
      author={Michael Hassid and Gabriel Synnaeve and Yossi Adi and Roy Schwartz},
      year={2025},
      eprint={2505.17813},
      archivePrefix={arXiv},
      primaryClass={cs.CL},
      url={https://arxiv.org/abs/2505.17813}, 
}

@misc{adaptthink,
      title={AdaptThink: Reasoning Models Can Learn When to Think}, 
      author={Jiajie Zhang and Nianyi Lin and Lei Hou and Ling Feng and Juanzi Li},
      year={2025},
      eprint={2505.13417},
      archivePrefix={arXiv},
      primaryClass={cs.CL},
      url={https://arxiv.org/abs/2505.13417}, 
}

@inproceedings{branchpredict,
  title={A study of branch prediction strategies},
  author={Smith, James E},
  booktitle={25 years of the international symposia on Computer architecture (selected papers)},
  pages={202--215},
  year={1998}
}

@misc{ircot,
      title={Interleaving Retrieval with Chain-of-Thought Reasoning for Knowledge-Intensive Multi-Step Questions}, 
      author={Harsh Trivedi and Niranjan Balasubramanian and Tushar Khot and Ashish Sabharwal},
      year={2023},
      eprint={2212.10509},
      archivePrefix={arXiv},
      primaryClass={cs.CL},
      url={https://arxiv.org/abs/2212.10509}, 
}

\end{document}